\pdfoutput=1
\documentclass[sigconf]{acmart}
\def\BibTeX{{\rm B\kern-.05em{\sc i\kern-.025em b}\kern-.08emT\kern-.1667em\lower.7ex\hbox{E}\kern-.125emX}}

\usepackage{balance}
\setcopyright{rightsretained}
\begin{document}

\fancyhead{}
 % do not delete this code.

% Updated according to the ACM Permission Release Form
%   - MM '19: 2019 ACM Multimedia Conference 201 fp631
\copyrightyear{2019} 
\acmYear{2019} 
\acmConference[MM '19]{Proceedings of the 27th ACM International Conference on Multimedia}{October 21--25, 2019}{Nice, France}
\acmBooktitle{Proceedings of the 27th ACM International Conference on Multimedia (MM '19), October 21--25, 2019, Nice, France}
\acmPrice{15.00}
\acmDOI{10.1145/3343031.3350988}
\acmISBN{978-1-4503-6889-6/19/10}

%
% The "title" command has an optional parameter, allowing the author to define a "short title" to be used in page headers.
\title[A Single-Shot Arbitrarily-Shaped Text Detector based on Multi-Task Learning]{A Single-Shot Arbitrarily-Shaped Text Detector based on Context Attended Multi-Task Learning}

%
% The "author" command and its associated commands are used to define the authors and their affiliations.
% Of note is the shared affiliation of the first two authors, and the "authornote" and "authornotemark" commands
% used to denote shared contribution to the research.

\author{Pengfei Wang}
\authornote{Equal contribution. This work is done when Pengfei Wang is an intern at Baidu Inc.}
\affiliation{%
  %\institution{Xidian University}
  \institution{School of Artificial Intelligence, Xidian University}
  %\streetaddress{2 S Taibai Road}
  %\city{Xi'an}
  %\state{Shaanxi}
  %\postcode{710071}
  %\country{China}
}
\email{pfwang@stu.xidian.edu.cn}

\author{Chengquan Zhang}
\authornotemark[1]
\affiliation{
  %\institution{Baidu Inc.}
  \institution{Department of Computer Vision Technology (VIS), Baidu Inc.}
  %\streetaddress{XXXX}
  %\city{Beijing}
  %\country{China}
  %\postcode{10000}
}
\email{zhangchengquan@baidu.com}

\author{Fei Qi}
\authornotemark[1]
\authornote{Corresponding author.}
\orcid{0000-0002-2161-1551}
\affiliation{%
  %\institution{Xidian University}
  \institution{School of Artificial Intelligence, Xidian University}
  %\streetaddress{2 S Taibai Road}
  %\city{Xi'an}
  %\state{Shaanxi}
  %\postcode{710071}
  %\country{China}
}
\email{fred.qi@ieee.org}

\author{Zuming Huang}
\affiliation{%
    \institution{Department of Computer Vision Technology (VIS), Baidu Inc.}
    % \city{Beijing}
    % \country{China}
}
\email{huangzuming@baidu.com}

\author{Mengyi En}
\affiliation{%
    \institution{Department of Computer Vision Technology (VIS), Baidu Inc.}
    % \city{Beijing}
    % \country{China}
}
\email{enmengyi@baidu.com}

\author{Junyu Han}
\affiliation{
  %\institution{Baidu Inc.}
  \institution{Department of Computer Vision Technology (VIS), Baidu Inc.}
}
\email{hanjunyu@baidu.com}

\author{Jingtuo Liu}
\affiliation{
  \institution{Department of Computer Vision Technology (VIS), Baidu Inc.}
}
\email{liujingtuo@baidu.com}

\author{Errui Ding}
\affiliation{
  \institution{Department of Computer Vision Technology (VIS), Baidu Inc.}
}
\email{dingerrui@baidu.com}

\author{Guangming Shi}
\orcid{0000-0003-2179-3292}
\affiliation{%
    %\institution{Xidian University}
    \institution{School of Artificial Intelligence, Xidian University}
    % \city{Xi'an}
    % \state{Shaanxi}
    % \country{China}
}
\email{gmshi@xidian.edu.cn}
%
% By default, the full list of authors will be used in the page headers. Often, this list is too long, and will overlap
% other information printed in the page headers. This command allows the author to define a more concise list
% of authors' names for this purpose.
\renewcommand{\shortauthors}{Wang and Zhang, et al.}

%
% The abstract is a short summary of the work to be presented in the article.
\begin{abstract}
Detecting scene text of arbitrary shapes has been a challenging task over the past years. In this paper, we propose a novel segmentation-based text detector, namely SAST, which employs a context attended multi-task learning framework based on a Fully Convolutional Network (FCN) to learn various geometric properties for the reconstruction of polygonal representation of text regions. Taking sequential characteristics of text into consideration, a Context Attention Block is introduced to capture long-range dependencies of pixel information to obtain a more reliable segmentation. In post-processing, a Point-to-Quad assignment method is proposed to cluster pixels into text instances by integrating both high-level object knowledge and low-level pixel information in a single shot. Moreover, the polygonal representation of arbitrarily-shaped text can be extracted with the proposed geometric properties much more effectively. Experiments on several benchmarks, including ICDAR2015, ICDAR2017-MLT, SCUT-CTW1500, and Total-Text, demonstrate that SAST achieves better or comparable performance in terms of accuracy. Furthermore, the proposed algorithm runs at 27.63 FPS on SCUT-CTW1500 with a Hmean of 81.0\% on a single NVIDIA Titan Xp graphics card, surpassing most of the existing segmentation-based methods. 

\end{abstract}

\begin{comment}
\begin{CCSXML}
<ccs2012>
<concept>
<concept_id>10010147.10010178.10010224</concept_id>
<concept_desc>Computing methodologies~Computer vision</concept_desc>
<concept_significance>300</concept_significance>
</concept>
<concept>
<concept_id>10010147.10010178.10010224.10010245.10010247</concept_id>
<concept_desc>Computing methodologies~Image segmentation</concept_desc>
<concept_significance>300</concept_significance>
</concept>
<concept>
<concept_id>10010147.10010178.10010224.10010225.10010227</concept_id>
<concept_desc>Computing methodologies~Scene understanding</concept_desc>
<concept_significance>100</concept_significance>
</concept>
</ccs2012>
\end{CCSXML}

\ccsdesc[300]{Computing methodologies~Computer vision}
\ccsdesc[300]{Computing methodologies~Image segmentation}
\ccsdesc[100]{Computing methodologies~Scene understanding}
\end{comment}

%
% Keywords. The author(s) should pick words that accurately describe the work being
% presented. Separate the keywords with commas.

% Detection
%Single Shot Detector
%

%
% A "teaser" image appears between the author and affiliation information and the body 
% of the document, and typically spans the page.

% \begin{comment}
%   \Description{Enjoying the baseball game from the third-base seats. Ichiro Suzuki preparing to bat.}
% \end{comment}

%
% This command processes the author and affiliation and title information and builds
% the first part of the formatted document.
\maketitle

\section{Introduction}

% Recently, scene text reading has attracted extensive attention in both academia and industry for its numerous applications, such as scene understanding, image and video retrieval, and robot navigation. As the prerequisite in the procedure of textual information extraction and understanding, text detection is thus of great importance. 
% Thanks to the surge of deep neural networks, various convolutional neural network (CNN) based methods has been proposed to detect scene text, continuously refreshing the performance records on standard benchmarks~\cite{karatzas2015icdar}.
% However, text detection in the wild is still a great challenging task due to the significantly  varied sizes, aspect ratios, orientations, languages, arbitrary shapes, and even the complex background under uncontrollable scenarios. In this paper, we seek an efficient and effective detector for text of arbitrary shapes.

Recently, scene text reading has attracted extensive attention in both academia and industry for its numerous applications, such as scene understanding, image and video retrieval, and robot navigation. As the prerequisite in textual information extraction and understanding, text detection is of great importance. 
Thanks to the surge of deep neural networks, various convolutional neural network (CNN) based methods have been proposed to detect scene text, continuously refreshing the performance records on standard benchmarks~\cite{karatzas2015icdar, nayef2017icdar2017, yuliang2017detecting, ch2017total}.
However, text detection in the wild is still a challenging task due to the significant variations in size, aspect ratios, orientations, languages, arbitrary shapes, and even the complex background. In this paper, we seek an effective and efficient detector for text of arbitrary shapes.

\begin{figure}
  \includegraphics[width=0.96 \linewidth]{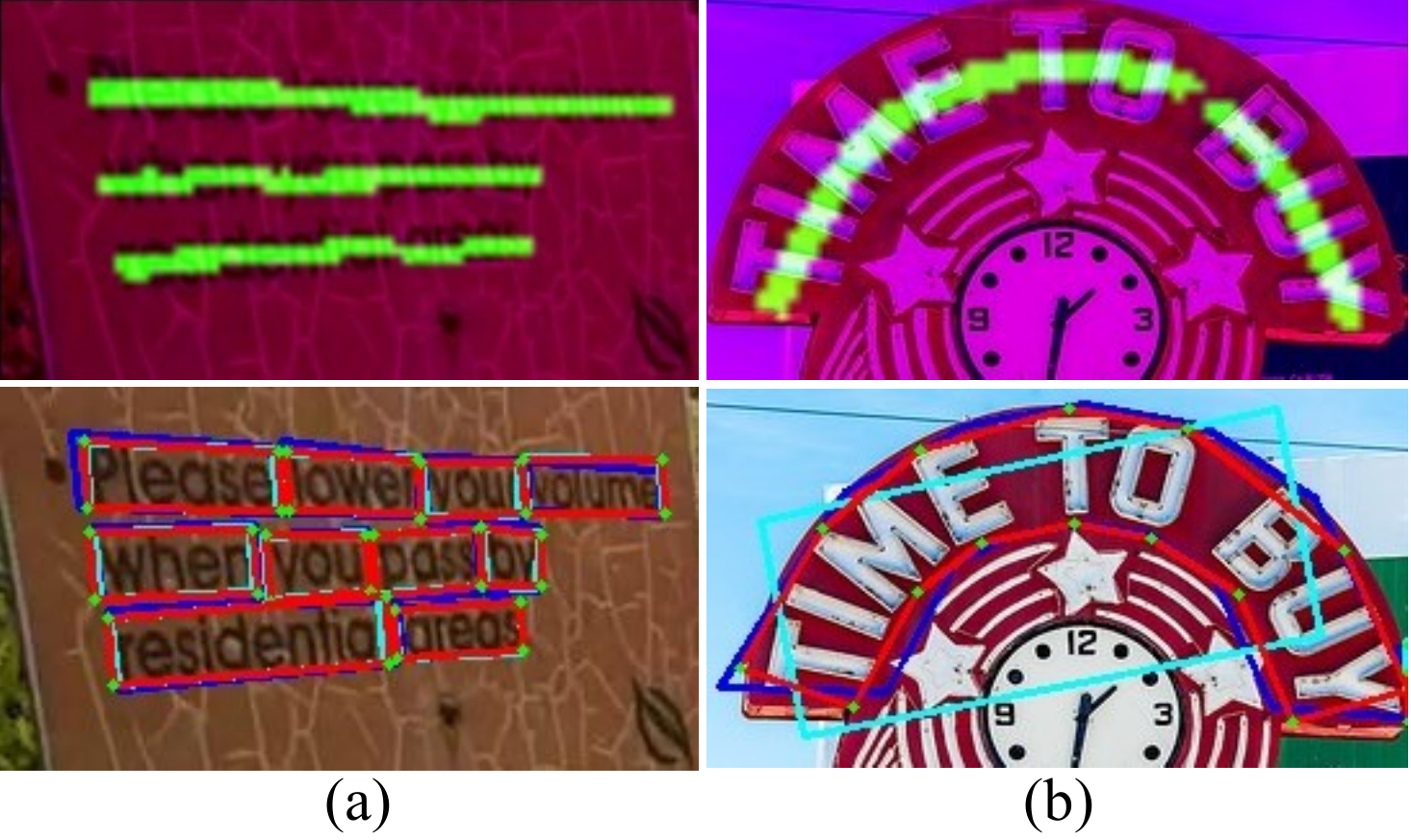}
  %\caption{Two common challenges for segmentation-based text detectors: (a) Adjacent text instances are difficult to separate; (b) The long text may break into fragments. The first row is the true label while the second row is response for text regions from segmentation branch. In the third row, cyan contours are the detection results from EAST~\cite{zhou2017east}, while red contours are from the proposed method, which alleviates the common challenges described above.} 
    \caption{Two common challenges for segmentation-based text detectors: (a) Adjacent text instances are difficult to be separated; (b) The long text may break into fragments. The first row is the response for text regions from segmentation branch. In the second row, cyan contours are the detection results from EAST~\cite{zhou2017east}, while red contours are from SAST.} 
  \label{fig:insight}
  %\vspace{-0.5cm}
\end{figure}

To detect arbitrarily-shaped text, especially those in curved form, some segmentation-based approaches~\cite{zhang2016multi, wu2017self, long2018textsnake,wang2019shape} formulated text detection as a semantic segmentation problem. They employ a fully convolutional network (FCN)~\cite{milletari2016v} to predict text regions, and apply several post-processing steps such as connected component analysis to extract final geometric representation of scene text. 
%The segmentation results of FCN and the post-processing steps depend mainly on local information.
Due to the lack of global context information, there are two common challenges for segmentation-based text detectors, as demonstrated in Fig.~\ref{fig:insight}, including: (1) Lying close to each other, text instances are difficult to be separated via semantic segmentation; (2) Long text instances tend to be fragmented easily, especially when character spacing is far or the background is complex, such as the effect of strong illumination.
In addition, most segmentation-based detectors have to output large-resolution prediction to precisely describe text contours, thus suffer from time-consuming and redundant post-processing steps.

Some instance segmentation methods~\cite{zheng2015conditional,he2017mask,fathi2017semantic} attempt to embed high-level object knowledge or non-local information into the network to alleviate the similar problems described above. Among them, Mask-RCNN~\cite{he2017mask}, a proposal-based segmentation method that cascades detection task (i.e., RPN~\cite{ren2015faster}) and segmentation task by RoIAlign~\cite{he2017mask}, has achieved better performance than those proposal-free methods by a large margin. Recently, some similar ideas~\cite{yao2018mask, huang2019mask, yang2018inceptext} have been introduced to settle the problem of detecting text of arbitrary shapes. However, they are all facing a common challenge that it takes much more time when the number of valid text proposals increases, due to the large number of overlapping computations in segmentation, especially in the case that valid proposals are dense. In contrast, our approach is based on a single-shot view and efficient multi-task mechanism.

Inspired by recent works~\cite{kirillov2017instancecut, uhrig2018box2pix, liu2018affinity} in general semantic instance segmentation, 
%we aim to design a segmentation-based method,  which integrates both the high-level object knowledge and low-level pixel information in a single shot and detects scene text of arbitrary shapes with high accuracy and efficiency.
we aim to design a segmentation-based \textbf{S}ingle-shot \textbf{A}rbitrarily-\textbf{S}haped  \textbf{T}ext detector (SAST), which integrates both the high-level object knowledge and low-level pixel information in a single shot and detects scene text of arbitrary shapes with high accuracy and efficiency.
Employing a FCN~\cite{milletari2016v} model, various geometric properties of text regions, including text center line (TCL), text border offset (TBO), text center offset (TCO), and text vertex offset (TVO), are designed to learn simultaneously under a multi-task learning formulation. In addition to skip connections, a Context Attention Block (CAB) is introduced into the architecture to aggregate contextual information for feature augmentation.
To address the problems illustrated in Fig.~\ref{fig:insight}, we propose a point-to-quad method for text instance segmentation, which assigns labels to pixels by combining high-level object knowledge from TVO and TCO maps. After clustering TCL map into text instances, more precise polygonal representations of arbitrarily-shaped text are then reconstructed based on TBO maps. 

Experiments on public datasets demonstrate that the proposed method achieves better or comparable performance in terms of both accuracy and efficiency. The contribution of this paper are three-fold:
% Notice gain, TCL, TBO, TCO and TVO maps are obtained by only single-shot prediction. 
% The experiments in public datasets demonstrate that the proposed method achieves better or comparable performance in terms of accuracy much more efficiently.
%The contribution of this paper are four-fold:
\begin{itemize}
\item[$\bullet$] We propose a single-shot text detector based on multi-task learning for text of arbitrary shapes including multi-oriented, multilingual, and curved scene text, which is efficient enough for some real-time applications.
%\item[$\bullet$] We propose a single-shot text detector named SAST for text of arbitrary shapes including long, multi-oriented and curved scene text, which is efficient enough for some real-time applications.
\item[$\bullet$] The Context Attention Block aggregates the contextual information to augment the feature representation without too much extra calculation cost.
\item[$\bullet$] The point-to-quad assignment is robust and effective to separate text instance and alleviate the problem of fragments, which is better than connected component analysis.
% \item[$\bullet$] Without bells and whistles, our approach achieves better or comparable performance on several public text detection benchmarks.
\end{itemize}
% ------------------- figure end ------------------

\section{Related Work}
%  Inspired by generic object detection and segmentation frameworks, the CNN based text detectors can be roughly divided into two categories: detection based method and segmentation based method. In this section, we will first review some representative methods proposed in generic instance-level segmentation tasks in Section 2.1 and Section 2.2.  and the segmentation based text detector is depicted in Section 2.2. A comprehensive review of recent scene text detectors can be found in ~\cite{ ye2015text, zhu2016scene}.

%Derived from generic object detection and segmentation frameworks, the CNN based text detectors can be roughly divided into two categories: detection-based methods and segmentation-based methods. 
In this section, we will review some representative segmentation-based and detection-based text detectors, as well as some recent progress in general semantic segmentation. A comprehensive review of recent scene text detectors can be found in~\cite{ ye2015text, zhu2016scene}.

\textbf{Segmentation-based Text Detectors.}
The greatest benefit of segmentation-based methods is the ability to detect both straight text and curved text in a unified manner. With FCN~\cite{milletari2016v}, the segmentation-based text detectors first classify text at pixel level, then followed by several post-processing to extract final geometric representation of scene text, so the performance of this kind of detectors is strongly affected by the robustness of segmentation results. In PixelLink~\cite{deng2018pixellink}, positive pixels are joined into text instances by predicted positive links, and the bounding boxes are extracted from segmentation result directly. TextSnake~\cite{long2018textsnake} proposes a novel presentation for arbitrarily-shaped text, and treats a text instance as a sequence of overlapping disks lying at text center line to describe the geometric properties of text instances of irregular shapes. PSENet~\cite{wang2019shape} shrinks the original text instance into various scales, and gradually expands the kernels to the text instances of complete shapes through a progressive scale expansion algorithm. The main challenge of FCN-based methods is separating text instances, which lie close to each other. The runtime of approaches above highly depends on the employed post-processing step, which often involves several pipelines and tends to be rather slow. 

\textbf{Detection-based Text Detectors.}
 Scene text is regarded as a special type of object, several methods~\cite{he2017deep,liao2017textboxes, liao2018rotation, ma2018arbitrary, zhou2017east, Zhang2019CVPR} are based on Faster R-CNN~\cite{ren2015faster}, SSD~\cite{Liu2016SSDSS} and DenseBox~\cite{huang2015densebox}, which generates text bounding boxes by regressing coordinates of boxes directly. TextBoxes~\cite{liao2017textboxes} and RRD~\cite{liao2018rotation} adopt SSD as a base detector and adjust the anchor ratios and convolution kernel size to handle variation of aspect ratios of text instances. He et al.~\cite{he2017deep} and EAST~\cite{zhou2017east} perform direct regression to determine vertex coordinates of quadrilateral text boundaries in a per-pixel manner without using anchors and proposals, and conduct the Non-Max Suppression (NMS) to get the final detection results. RRPN~\cite{ma2018arbitrary} generates inclined proposals with text orientation angle information and propose Rotation Region-of-Interest (RRoI) pooling layer to detect arbitrary-oriented text. Limited by the receptive field of CNNs and the relatively simple representations like rectangle bounding box or quadrangle adopted to describe text, detection-based methods may fall short when dealing with more challenging text instances, such as extremely long text and arbitrarily-shaped text.

\textbf{General Instance Segmentation.}
Instance segmentation is a challenging task, which involves both segmentation and classification tasks. 
%Some frameworks generate object proposals at first, and employ additional networks to achieve a pixel-accurate object mask as final segmentation. 
The most recent and successful two-stage representative is Mask R-CNN~\cite{he2017mask}, which achieves amazing results on public benchmarks, but requires relatively long execution time due to the per-proposal computation and its deep stem network. Other frameworks rely mostly on pixel-features generated by a single FCN forward pass, and employ post-processing like graphical models, template matching, or pixel embedding to cluster pixels belonging to the same instance. More specifically, Non-local Networks~\cite{wang2018non} utilizes a self-attention~\cite{vaswani2017attention} mechanism to enable a pixel-feature to perceive features from all the other positions, while the CCNet~\cite{huang2018ccnet} harvests the contextual information from all pixels more efficiently by stacking two criss-cross attention modules, which augments the feature representation a lot. In post-processing step, Liu et al.~\cite{liu2018affinity} present a pixel affinity scheme and cluster pixels into instances with a simple yet effective graph merge algorithm. Instance-Cut~\cite{kirillov2017instancecut} and the work of \cite{yu2018learning} predict object boundaries intentionally to facilitate the separation of object instances.

Our method, SAST, employs a FCN-based framework to predict TCL, TCO, TVO, and TBO maps in parallel. With the efficient CAB and point-to-quad assignment, where the high-level object knowledge is combined with low-level pixel information, SAST can detect text of arbitrary shapes with high accuracy and efficiency.
% ------------------- figure begin ----------------
\begin{figure}
  \includegraphics[width=\linewidth]{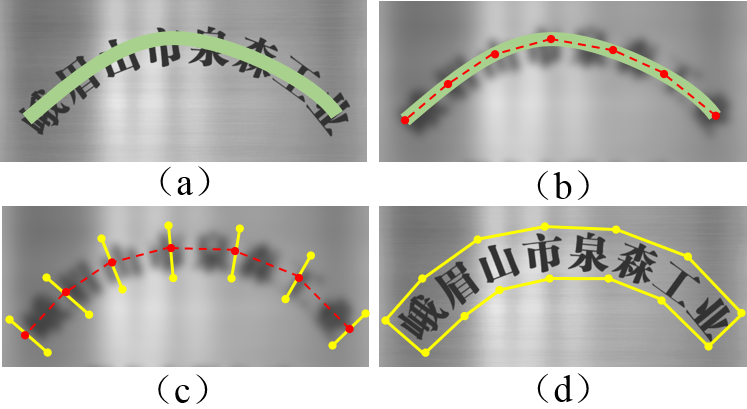}
  \caption{Arbitrary Shape Representation: a) The text line in TCL map; b) Sample adaptive number of points in the  line; c) Calculate the corresponding border point pairs with TBO map; d) Link all the border points as final representation. }
  \label{fig:ase}
  %\vspace{-0.5cm}
\end{figure}
% ------------------- figure end ------------------

\begin{figure*}
  \includegraphics[width=\linewidth]{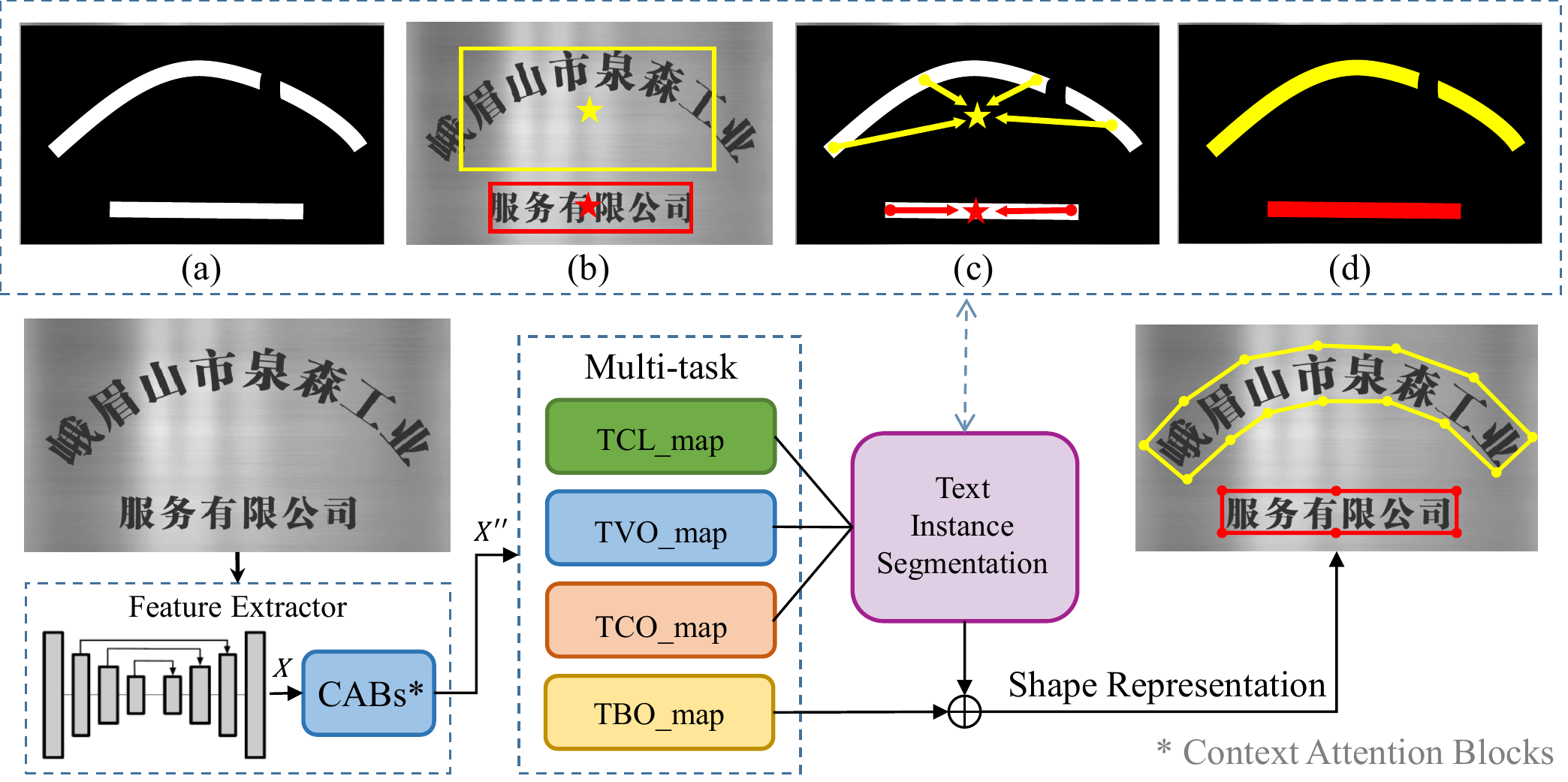}
  \caption{The pipeline of proposed method: 1) Extract feature from input image, and learn TCL, TBO, TCO, TVO maps as a multi-task problem; 2) Achieve instance segmentation by Text Instance Segmentation Module, and the mechanism of point-to-quad assignment is illustrated in c; 3) Restore polygonal representation of text instances of arbitrary shapes.}
  \label{fig:pipline}
  %\vspace{0.5cm}
\end{figure*}
% ------------------- figure end ------------------

% we define TVO to predict the vertices of bounding quadrangle of arbitrary shape text region, and apply Non-Maximum Suppression (NMS) to get the detection results as high-level object knowledge. 
\section{METHODOLOGY}
% In this section, we describe our SAST framework for detection scene text of arbitrary shapes, which is supposed to run in the real-time speed, in details.
In this section, we will describe our SAST framework for detecting scene text of arbitrary shapes in details.

\subsection{Arbitrary Shape Representation}

% The expression of quadrangle fails to precisely describe the text instances of arbitrary shapes, and the contours of text instances are extracted as the final shape expression for some segmentation-based text detector. To avoid sticking with each other, PSENet \cite{wang2019shape} adopts a progressive scale expansion algorithm to expands the kernels with minimal scales to get final contours of text instances. Moreover, TextSnake~\cite{long2018textsnake} adopts a sequence of overlapping disks lying at the centre line to describe the geometric properties of text instances of irregular shapes, of which the post-procession is complex and tended to be slow.
% ------------------- figure begin ----------------
The bounding boxes, rotated rectangles, and quadrilaterals are used as classical representations in most detection-based text detectors, which fails to precisely describe the text instances of arbitrary shapes, as shown in Fig.~\ref{fig:insight} (b). 
The segmentation-based methods formulate the detection of arbitrarily-shaped text as a binary segmentation problem. Most of them directly extracted the contours of instance mask as the representation of text, which is easily affected by the completeness and consistency of segmentation. However, PSENet~\cite{wang2019shape} and TextSnake~\cite{long2018textsnake} attempted to progressively reconstruct the polygonal representation of detected text based on a shrunk text region, of which the post-processing is complex and tended to be slow. Inspired by those efforts, we aim to design an effective method for arbitrarily-shaped text representation.
 
% The horizontal bounding boxes, rotated rectangles, and quadrilaterals are used as classical text representations in most detection-based text detectors, which fails to precisely describe the text instances of arbitrary shapes. The segmentation-based methods formulate the detection of arbitrarily-shaped text as a segmentation problem. However, The naive binary text mask representation of text, which is easy to stick together, is redundant and not suitable for recognition task. The shrunk text region is commonly learned to make it feasible to separate text instances that are close to each other. It comes with a question that how to reconstruct the polygonal representation of detected text region efficiently. PSENet \cite{wang2019shape} adopts a progressive scale expansion algorithm to expands the kernels several times with minimal scales to get the contours of text instances as final polygonal representation, of which the post-procession is complex and tended to be slow.

In this paper, we extract the center line of text region (TCL map) and reconstruct the precise shape representation of text instances with a regressed geometry property, i.e. TBO, which indicates the offset between each pixel in TCL map and corresponding point pair in upper and lower edge of its text region. More specifically, as depicted in Fig.~\ref{fig:ase}, the representation strategy consists of two steps: text center point sampling and border point extraction. Firstly, we sample $n$ points at equidistance intervals from left to right on the center line region of text instance. By taking a further operation, we can determine the corresponding border point pairs based on the sampled center line point with the information provided by TBO maps in the same location. By linking all the border points clockwise, we can obtain a complete text polygon representation. Instead of setting $n$ to a fixed number, we assign it by the ratio of center line length to the average of length of border offset pairs adaptively. Several experiments on curved text datasets prove that our method is efficient and flexible for arbitrarily-shaped text instances. 

% Considering the  we set n to 7 in the curved text detection according to the label definition in the SCUT-CTW1500~\cite{yuliang2017detecting, ch2017total}, and to 2 when dealing with text detection in such benchmarks~\cite{karatzas2015icdar, nayef2017icdar2017, shi2017icdar2017} labeled with quadrangle annotations.

\subsection{Pipeline}

The network architecture of FCN-based text detectors are limited to the local receptive fields and short-range contextual information, and makes it struggling to segment some challenging text instances. Thus, we design a \textbf{Context Attention Block} to integrate the long-range dependencies of pixels to obtain a more representative feature. As a substitute for the Connected Component Analysis, we also propose \textbf{Point-to-Quad Assignment} to cluster the pixels in TCL map into text instances, where we use TCL and TVO maps to restore the minimum quadrilateral bounding boxes of text instances as high-level information.

% which is based on TVO and TCO maps and integrates both high-level object knowledge and low-level pixel information. 

An overview of our framework is depicted in Fig.~\ref{fig:pipline}. It consists of three parts, including a stem network, multi-task branches, and a post-processing part. The stem network is based on ResNet-50~\cite{he2016deep} with FPN~\cite{lin2017feature} and CABs to produce context-enhanced representation. The TCL, TCO, TVO, and TBO maps are predicted for each text region as a multi-task problem. In the post-processing, we segment text instances by point-to-quad assignment. Concretely, similar to EAST~\cite{zhou2017east}, the TVO map regresses the four vertices of bounding quadrangle of text region directly, and the detection results is considered as high-level object knowledge. For each pixel in TCL map, a corresponding offset vector from TCO map will point to a low-level center which the pixel belongs to. Computing the distance between lower-level center and high-level object centers of the detected bounding quadrangle, pixels in the TCL map will be grouped into several text instances. In contrast to the connected component analysis, it takes high-level object knowledge into account, and is proved to be more efficient. More details about the mechanism of point-to-quad assignment will be discussed in this Section 3.4. We sample a adaptive number of points in the center line of each text instance, calculate corresponding points in upper and lower borders with the help of TBO map, and reconstruct the representation of arbitrarily-shaped scene text finally.

% by clustering TCL responses with the help of geometric information in TCO and TVO maps.

% In the post-processing step, the high-level object knowledge from TVO map and the pixel-level information from TCO map will be integrated to segment TCL map into text instances . Then, the final geometric representation of arbitrarily-shaped text instances can be reconstructed with the boundary offset information from the TBO map.

% Concretely, TCL, as a score map for text or non-text, is predicted by the segmentation branch. Different from previous works that adopt connected component analysis in post-processing, The proposed post-processing procedure can be roughly divided into there phases. 

% 1) Detection Phase.  2) Text Instance Segmentation Phase.  In contrast to the connected component analysis, it takes high-level object knowledge into account, instead of only the local information of neighborhood pixels, and proved to be more efficient.  3) Shape Representation Phase. We sample a adaptive number of points in the center line of each text instance, calculate corresponding points in upper and lower border with the regressed TBO map, and reconstruct the geometric representation of arbitrarily-shaped scene text.

\subsection{Network Architecture}

In this paper, we employ ResNet-50 as the backbone network with the additional fully-connected layers removed. With different levels of feature map from the stem network gradually merged three-times in the FPN manner, a fused feature map $X$ is produced at $1/4$ size of the input images. We serially stack two CABs behind to capture rich contextual information. Adding four branches behind the context-enhanced feature maps $X''$, the TCL and other geometric maps are predicted in parallel, where we adopt a $1\times 1$ convolution layer with the number of output channel set to \{1, 2, 8, 4\} for TCL, TCO, TVO, and TBO map respectively. It is worth mentioning that all the output channels of convolution layers in the FPN is set to 128 directly, regardless of whether the kernel size is 1 or 3.

\begin{figure}
  \includegraphics[width=\linewidth]{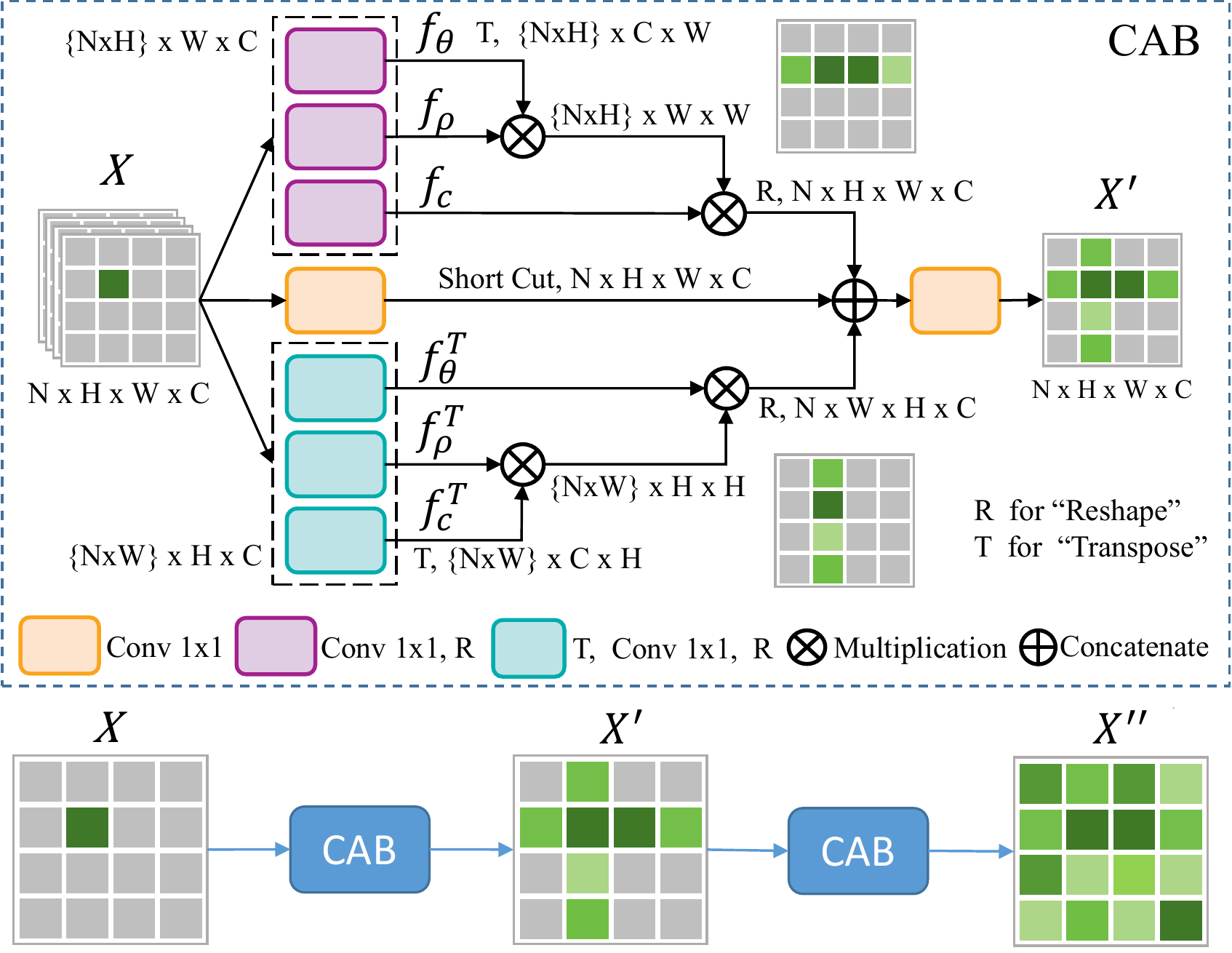}
  \caption{ Context Attention Blocks: a single CAB module aggregates pixel-wise contextual information both horizontally and vertically, and long-range dependencies from all pixels can be captured by serially connecting two CABs.}
  \label{fig:cab_module}
  %%\vspace{0.5cm}
\end{figure}

\textbf{Context Attention Block.} The segmentation results of FCN and the post-processing steps depend mainly on local information. The proposed CAB utilizes a self-attention mechanism~\cite{vaswani2017attention} to aggregate the contextual information to augment the feature representation, of which the details is demonstrated in Fig.~\ref{fig:cab_module}. 
In order to alleviate the huge computational overhead caused by direct use of self-attention, CAB only considers the similarity between each location in feature map and others in the same horizontal or vertical column.
The feature map  $X$ is the output of ResNet-50 backbone which is in size of $N \times H \times W \times C$. 
To collect contextual information horizontally, we adopt three convolution layers behind $X$ in parallel to get \{ $f_\theta$, $f_\phi$, $f_g$ \} and reshape them into  $\{N \times H\}  \times W  \times C$, then multiply $f_\phi$ by the transpose of $f_\theta$ to get an attention map of size $\{N \times H\} \times W \times W$, which is activated by a Sigmoid function. A horizontal contextual information enhanced feature, which is resized to $N \times H \times W \times C$ finally, is integrated by multiplication of $f_g$ and the attention map. 
It is slightly different to get the vertical contextual information that \{ $f_\theta$,$f_\phi$,$f_g$ \} is transposed to  $\{N \times H\}  \times C  \times W$ at the beginning, as shown in cyan boxes in Fig.~\ref{fig:cab_module}. Meanwhile, a short-cut path is used to preserve local features. Concatenating the horizontal contextual map, vertical contextual map,  and short-cut map together and reducing channel number of $ X' $  with a $1 \times 1$ convolutional layer, the CAB aggregates long-range pixel-wise contextual information in both horizontal and vertical directions. Besides, the convolutional layers denoted by purple and cyan boxes share weights. By serially connecting two CABs, each pixel can finally capture long-range dependencies from all pixels, as depicted in the bottom of Fig.~\ref{fig:cab_module}, leading to a more powerful context-enhanced feature map $X''$, which also helps alleviate the problems caused by the limited receptive field when dealing with more challenging text instances, such as long text.

% ------------------- figure begin ----------------
% ------------------- figure end ------------------

% ------------------- figure begin ----------------
% \begin{figure*}
%   \ing
%   \includegraphics[width=\linewidth]{image/network.png}
%   \caption{The proposed architecture}
%   \label{fig:network}
% \end{figure*}
% % ------------------- figure end ------------------

%\subsection{Point-to-Quad Assignment}
\subsection{Text Instance Segmentation}
For most proposal free text detector of arbitrary shapes, the morphological post-processing such as connected component analysis are adopted to achieve text instance segmentation, which do not explicitly incorporate high-level object knowledge and easily fail to detect complex scene text. In this section, we describe how to generate an text instance semantic segmentation with TCL, TCO and TVO maps with high-level object information.
%For most proposal free text detector of arbitrary shape, the morphological post-processing such as connected component analysis are adopted to achieve text instance segmentation, which do not explicitly incorporate high-level object knowledge and tend to be a time-consuming task. In this section, we describe how to generate an text instance semantic segmentation with TCL, TCO and TVO maps with high-level object information.

\textbf{Point-to-Quad Assignment.}
As depicted in Fig.~\ref{fig:pipline}, the first step of text instance segmentation is detecting candidate text quadrangles based on TCL and TVO maps. 
Similar to EAST~\cite{zhou2017east}, we binarize the TCL map, whose pixel values are in the range of [0, 1], with a given threshold, and restore the corresponding quadrangle bounding boxes with the four vertex offsets provided by TVO map. Of course, NMS is adopted to suppress overlapping candidates.
The final quadrangle candidates shown in Fig.~\ref{fig:pipline} (b) can be considered to depend on high-level knowledge.
The second and last step in text instance segmentation is clustering the responses of text region in the binarized TCL map into text instances.
As Fig.~\ref{fig:pipline} (c) shows, the TCO map is a pixel-wise prediction of offset vectors pointing to the center of bounding boxes which the pixels in the TCL map should belong to. 
With a strong assumption that pixels in TCL map belonging to the same text instance should point to the same object-level center, we cluster TCL map into several text instances by assigning the response pixel to the quadrangle boxes generated in the first step.
Moreover, we do not care about whether predicted boxes in the first step are fully bounding text region in the input image, and the pixels outside of the predicted box will be mostly assigned to the corresponding text instances.
Integrated with high-level object knowledge and low-level pixel information, the proposed post-processing clusters each pixel in TCL map to its best matching text instance efficiently, and can help to not only separate text instances that are close to each other, but also alleviate fragments when dealing with extremely long text.

\subsection{Label Generation and Training Objectives}
\begin{figure}
  \includegraphics[width=\linewidth]{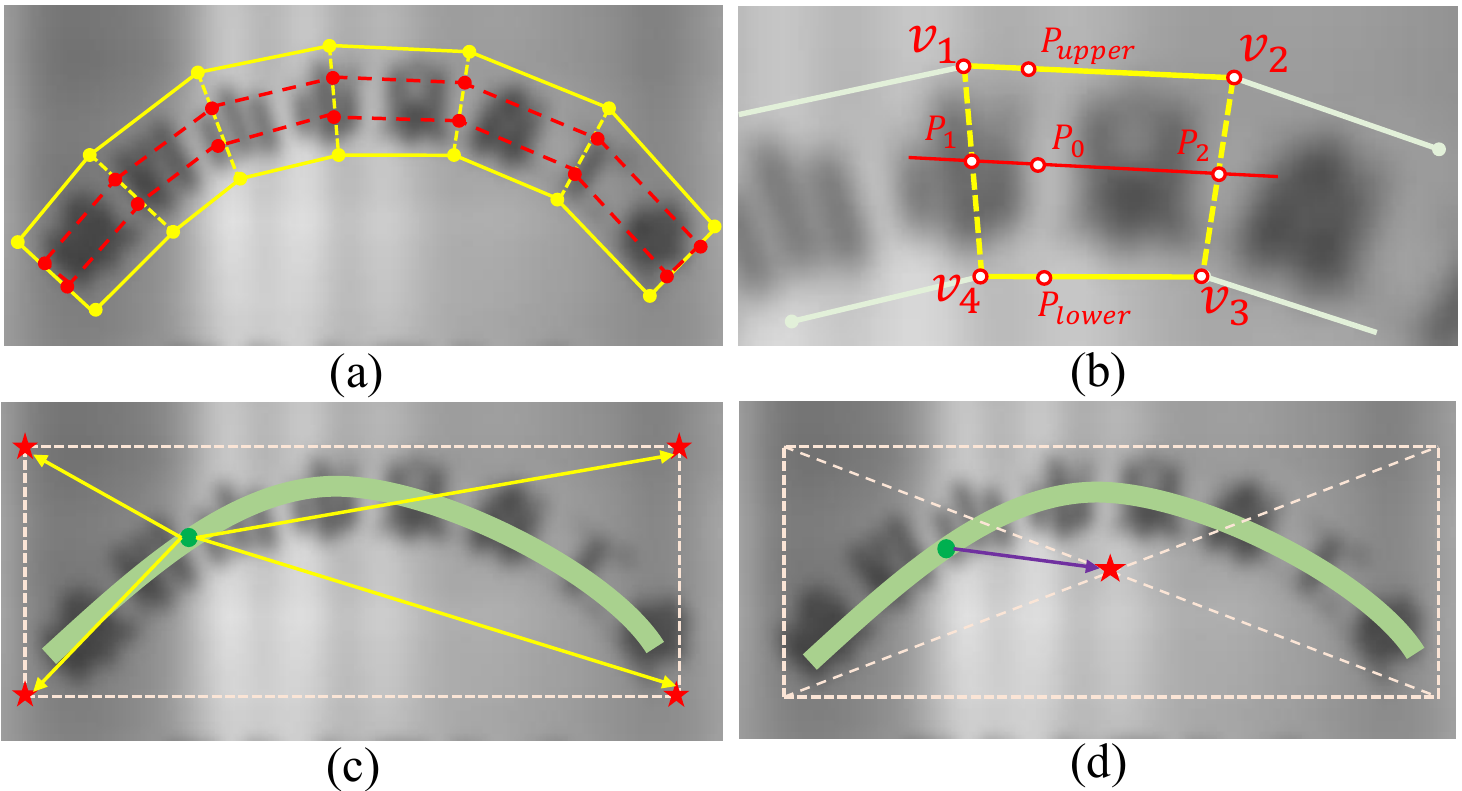}
  \caption{Label Generation: (a) Text center region of a curved text is annotated in red; (b) The generation of TBO map; The four vertices (red stars in c) and center point(red star in d) of bounding box, to which the TVO and TCO map refer.  }
  \label{fig:label_gen}
  %%\vspace{0.5cm}
\end{figure}

In this part, the generation of TCL, TCO, TVO, and TBO maps will be discussed. TCL is the shrunk version of text region, and it is an one channel segmentation map for text/non-text. 
The other label maps such as TCO, TVO, and TBO, are per-pixel offsets with reference to those pixels in TCL map. 
For each text instance, we calculate the center and four vertices of the minimum enclosing quadrangle from its annotation polygon, as depicted in Fig.~\ref{fig:label_gen} (c) (d). TCO map is the offset between pixels in TCL map and the center of bounding box, while TVO is the offset between the four vertices of bounding box and pixels in TCL map. 
Hence, the channel numbers of TCO and TVO maps are 2 and 8, respectively, because each point pair requires two channels to represent the offsets \{$\Delta x_i$, $\Delta y_i$\}.
Meanwhile, TBO determines the upper and lower boundaries of text instances in TCL map, thus it is a four channel offset map.
%In this part, the detail of TCL, TCO, TVO, and TBO maps will be discussed. TCL is the shrunk version of text region, and it is an one channel segmentation map as classification for text/non-text. The other label maps such as TCO, TVO, and TBO, is per-pixel offset with reference to those pixels in TCL map. For each text region, we extract the center and vertices of bounding rectangle from its annotation polygon, as depicted in Fig. \ref{fig:label_gen}(b)(c). TCO map is the offset between pixels in TCL map and the center of bounding box, while TVO is the offset between the four vertices of bounding box and pixels in TCL map. The channel number of TCO and TVO map is \{2, 8\} indicating the corresponding offset \{$\Delta x_1$, $\Delta y_1$\} and \{$\Delta x_1$, $\Delta y_1$, $\Delta x_2$, $\Delta y_2$, $\Delta x_3$, $\Delta y_3$, $\Delta x_4$, $\Delta y_4$\} respectively. Meanwhile, TBO is of great importance for arbitrary shape expression, which determines the upper and lower boundaries of text instances in TCL map. Thus, The TBO map is four channel offset map \{$\Delta x_1$, $\Delta y_1$, $\Delta x_2$, $\Delta y_2$ \} with reference to the corresponding boundary point pair.

% ------------------- figure begin ----------------
% ------------------- figure end ------------------

Here are more details about the generation of a TBO map. For a quadrangle text annotation with vertices \{$V_1$, $V_2$, $V_3$, $V_4$ \} in clock-wise and $V_1$ is the top left vertex, as shown in Fig.~\ref{fig:label_gen} (b), the generation of TBO mainly contains two steps: first, we find a corresponding point pair on the top and bottom boundaries for each point in TCL map, than calculate the corresponding offset pair. With average slope of the upper side and lower side of quadrangle, the line cross a point $P_0$  in TCL map can be determined. And it is easy to directly calculate the intersection points \{$P_1$, $P_2$ \} of the line in the left and right edges of bounding quadrangle with algebraic methods. A pair of corresponding points \{$P_{upper}$, $P_{lower}$ \} for $P_0$ can be determined from: 
$$ \frac{P_0 - P_1}{P_2 - P_1} = \frac{P_{upper} - V_1 }{V_2 - V_1} = \frac{P_{lower} - V_4}{V_3 - V_4}. $$ 
In the second step, the offsets between $P_0$ and \{$P_{upper}$, $P_{lower}$ \} can be easily determined. Polygons of more than four vertices are treated as a series of quadrangles connected together, and TBO of polygons can be generated gradually from quadrangles as described before. For non-TCL pixels, their corresponding geometry attributes are set to 0 for convenience.

At the stage of training, the whole network is trained in an end-to-end manner, and the loss of the model can be formulated as:
$$L_{total} = {\lambda}_1 L_{tcl} + {\lambda}_2 L_{tco} + {\lambda}_3 L_{tvo} +{\lambda}_4 L_{tbo}, $$
where $L_{tcl}$, $L_{tco}$,$L_{tvo}$ and $L_{tbo}$ represent the loss of TCL, TCO, TVO, and TBO maps, and the first one is the binary segmentation loss while the other are regression loss.
% we train the segmentation branch by minimizing the Dice loss~\cite{milletari2016v}, and the Smooth $L_1$ loss with the hyper-parameter $\alpha$ set to 1.0 is adopted for regression loss. The loss weight  ${\lambda}_1$, ${\lambda}_2$, ${\lambda}_3$, and ${\lambda}_4$ are set to $\{1.0, 0.5, 0.5, 1.0\}$ in our experiments.

% new added
% We train segmentation branch by minimizing the Dice loss~\cite{milletari2016v}, and the Smooth $L_1$ loss with the hyper-parameter $\alpha$ set to 1.0 is adopted for regression loss. The loss weights ${\lambda}_1$, ${\lambda}_2$, ${\lambda}_3$, and ${\lambda}_4$ are a tradeoff between four tasks which are equally important in this work, so we determine a set of values \{1.0, 0.5, 0.5, 1.0\} by making the four loss gradient norms close in back-propagation. 

% new updated
We train segmentation branch by minimizing the Dice loss~\cite{milletari2016v}, and the Smooth $L_1$ loss~\cite{fastrcnn} is adopted for regression loss. The loss weights ${\lambda}_1$, ${\lambda}_2$, ${\lambda}_3$, and ${\lambda}_4$ are a tradeoff between four tasks which are equally important in this work, so we determine a set of values \{1.0, 0.5, 0.5, 1.0\} by making the four loss gradient norms close in back-propagation.

\section{Experiments}

To compare the effectiveness of SAST with existing methods, we perform thorough experiments on four public text detection datasets, i.e., ICDAR 2015, ICDAR2017-MLT, SCUT-CTW1500 and Total-Text. 

\subsection{Datasets}
The datasets used for the experiments in this paper are briefly introduced below.

\textbf{SynthText.} The SynthText dataset ~\cite{gupta2016synthetic} is composed of 800,000 natural images, on which text in random colors, fonts, scales, and orientations is rendered carefully to have a realistic look. We use the dataset with word-level labels to pre-train our model.

\textbf{ICDAR 2015.} The ICDAR 2015 dataset~\cite{karatzas2015icdar} is collected for the ICDAR 2015 Robust Reading Competition, with 1,000 natural images for training and 500 for testing. The images are acquired using Google Glass and the text accidentally appear in the scene. All the text instances annotated with word-level quadrangles.

\textbf{ICDAR2017-MLT.} The ICDAR2017-MLT ~\cite{nayef2017icdar2017} is a large scale multi-lingual text dataset, which includes 7,200 training images, 1,800 validation images and 9,000 test images. The dataset consists of multi-oriented and multi-lingual aspects of scene text. The text regions in ICDAR2017-MLT are also annotated by quadrangles.

% \textbf{ICDAR2017-RCTW.} The ICDAR2017-RCTW ~\cite{shi2017icdar2017}   comprises 8034 training images and 4229 test images with scene texts printed in either Chinese or English. The images are captured from different sources including street views, posters, screen-shot, etc. Multi-oriented words and text lines are annotated using quadrangles.

\textbf{SCUT-CTW1500.} The SCUT-CTW1500~\cite{yuliang2017detecting} is a challenging dataset for curved text detection. It consists of 1,000 training images and 500 test images, and text instances are largely in English and Chinese. Different from traditional datasets, the text instances in SCUT-CTW1500 are labelled by polygons with 14 vertices.

\textbf{Total-Text.} The Total-Text~\cite{ch2017total} is another curved text benchmark, which consists of 1,255 training images and 300 testing images with more than 3 different text orientations: Horizontal, Multi-Oriented, and Curved. The annotations are labelled in word-level.

% \textbf{Evaluation Metrics.} The performance of SAST in ICDAR2015, ICDAR2017-MLT and ICDAR2017-RCTW, SCUT-CTW1500 and Total-Text is evaluated by using the evaluation protocol in ~\cite{karatzas2015icdar, nayef2017icdar2017, shi2017icdar2017,yuliang2017detecting, ch2017total}, respectively.

\textbf{Evaluation Metrics.} The performance on ICDAR2015, Total-Text, SCUT-CTW1500, and ICDAR2017-MLT is evaluated using the protocols provided in \cite{karatzas2015icdar, ch2017total, yuliang2017detecting, nayef2017icdar2017}, respectively.

\subsection{Implementation Details}
\textbf{Training.} ResNet-50 is used as the network backbone with pre-trained weight on ImageNet~\cite{deng2009imagenet}. The skip-connection is in FPN fashion with output channel numbers of the convolutional layers set to 128 and the final output is at 1/4 size of input images. All upsample operators are the bilinear interpolation and the classification branch is activated with sigmoid while the regression branches, i.e. TCO, TVO, and TBO maps, is the output of the last convolution layer directly. The training process is divided into two steps, i.e., the warming-up and fine-tuning steps. In the warming-up step, we apply Adam optimizer to train our model with learning rate 1e-4, and the learning rate decay factor is 0.94 on the SynthText dataset. In the fine-tuning step, the learning rate is re-initiated to 1e-4 and the model is tuned on ICDAR 2015, ICDAR2017-MLT, SCUT-CTW1500 and Total-Text.

All the experiments are performed on a workstation with the following configuration, CPU: Intel(R) Xeon(R) CPU E5-2620 v2 @ 2.10GHz x16; GPU: NVIDIA TITAN Xp $\times 4$; RAM: 64GB. During the training time, we set the batch size to 8 per GPU in parallel. 
%All the experiments are performed on a standard workstation with the following configuration, CPU: Intel(R) Xeon(R) CPU E5-2620 v2 @ 2.10GHz x16; GPU: Tesla K40m x4; RAM: 64GB. During the training time, we set the batch size to 8 on 4 GPUs in parallel. 

\textbf{Data Augmentation.} We randomly crop the text image regions, then resize and pad them to $512 \times 512$. Specially for curved polygon labeled datasets, we crop images without crossing text instances to avoid the destruction of polygon annotations. The cropped image regions will be rotated randomly in 4 directions ($0^\circ$, $90^\circ$, $180^\circ$, and $270^\circ$) and standardized by subtracting the RGB mean value of ImageNet dataset. The text region, which is marked as "DO NOT CARE" or its minimum length of edges is less than 8 pixels, will be ignored in the training process.

% new add
\textbf{Testing.} In inference phase, unless otherwise stated, we set the longer side to 1536 for single-scale testing, and to 512, 768, 1536, and 2048 for multi-scale testing, while keeping the aspect ratio unchanged. A specified range is assigned to each testing scale and detections from different scales are combined using NMS, which is inspired by SNIP\cite{singh2018analysis}.

%\textbf{Testing.} All the experiments are performed on a standard workstation with the following configuration, CPU: Intel(R) Xeon(R) CPU E5-2620 v2 @ 2.10GHz x16; GPU: Tesla K40m x4; RAM: 160GB. During the training time, we set the batch size to 8 on 4 GPUs in parallel. In inference phase, the batch size is set to 1 on 1 GPU.
% , and run testing on workstations equipped with computing processor graphic cards with different compute capability.

\subsection{Ablation Study}
We conduct several ablation experiments to analyze SAST. The details are discussed as follows.

\textbf{The Effectiveness of TBO, TCO and TVO.} To verify the efficiency of Text Instance Segmentation Module (TVO and TCO maps) and Arbitrary Shape Representation Module (TBO map) in SAST, we conduct several experiments with the following configurations: 1) TCL + CC + Expand: It is a naive way to predict center region of text, use connected component analysis to achieve text instance segmentation, and expand the contours of connected components by a shrinking rate as the final text geometric representation. 2) TCL + CC + TBO: Instead of expending the contours directly, we reconstruct the precise polygon of a text instance with Arbitrary Shape Representation Module. 3) TCL + TVO + TCO +TBO: As a substitute for connected component analysis, we use the method of point-to-quad assignment in Text Instance Segmentation Module, which is supposed to incorporate high-level object knowledge and low-level information, and assign each pixel on the TCL map to its best matching text instances. The efficiency of the proposed method is demonstrated on SCUT-CTW1500, as shown in Tab.~\ref{tab:ab_main}. It surpasses the first two methods by 21.75\% and 1.46\% in Hmean, respectively. Meanwhile, the proposed point-to-quad assignment cost almost the same time as connected component analysis.

\textbf{The Trade-off between Speed and Accuracy.}
There is a trade-off  between speed and accuracy, and the mainstream segmentation methods maintain high resolution, which is usually in the same size of input image, to achieve a better result at a correspondingly high cost in time. Several experiments are made on SCUT-CTW1500 benchmark, We compare the performance with different resolution of output, i.e., \{1, 1/2, 1/4, 1/8 \}, and find a rational trade-off between speed and accuracy at the 1/4 scale of input images. The detail configuration and results are shown in Tab.~\ref{tab:ab_tradeoff}. Note that the feature extractor in those experiments is not equipped with Context Attention Block.

% --------------------- table begin -----------------------
\begin{table}
  \caption{Ablation study for the effectiveness of TBO, TCO, and TVO in the proposed method. }
  \label{tab:ab_main}
  \begin{tabular}{cc@{ }c@{ }c@{ }c}
    \toprule
    Method &Recall & Precision & Hmean & T (ms) \\
    \midrule
    TCL + CC + expand  &55.51	&61.55	&58.37	&--\\
    TCL + CC + TBO  &74.65	&83.13	&78.66	&\textbf{5.84} \\
    TCL + TVO + TCO + TBO  &\textbf{76.69}	&\textbf{83.89}	&\textbf{80.12}	&6.26\\
    \bottomrule
  \end{tabular}
\end{table}
% ---------------- table end ---------------------------

% --------------------- table begin -----------------------
\begin{table}

  \caption{Ablation study for the trade-off between speed and accuracy on SCUT-CTW1500. }
  \label{tab:ab_tradeoff}
  \begin{tabular}{ccccc}
    \toprule
     Method &Recall&Precision&Hmean & FPS\\
    \midrule
    1s &76.34 &86.23 &80.98  &10.54 \\
    2s &70.86 &86.30 &77.82  &21.68 \\
    4s &76.69 &83.86 &80.12  &30.21 \\
    8s &71.54 &80.67 &75.83  &38.05 \\
  \bottomrule
\end{tabular}
% \vspace{-0.2cm}
% \vspace{-0.4cm}
\end{table}

% ---------------- table end ---------------------------

\textbf{The Effectiveness of Context Attention Blocks.} We introduce the CABs into the network architecture to capture long-range dependencies of pixel information. We conduct two experiments on SCUT-CTW1500 by replacing the CABs with several convolutional layers stacked together as the baseline experiment, which has almost the same number of trainable variables. The input size of image is $512 \times 512$, and the output of images is at 1/4 of input size. Tab.~\ref{tab:attention} demonstrates the performance and speed of both experiments. The experiment with CABs achieves 80.97\% in Hmean at a speed of 27.63 FPS, which exceeds the baseline by 0.85\% in Hmean at a bit slower frame rate.
%\textbf{The Effectiveness of Context Attention Block.} We introduce the CAB into the network architecture to capture long-range dependencies of pixel information to obtain a more reliable representation. We conduct two experiments on SCUT-CTW1500 validation set by replacing the CABs with several convolutional layers stacked together as the baseline experiment, which has almost the same number of trainable variables. The input size of image is $512 \times 512$, and the output of images is at 1/4 of input size. Tab. \ref{tab:attention} demonstrates the performance and speed information of the two experiments, and the experiment with CABs achieves 80.97 in hmean at a speed of 27.63 FPS, which is exceeds by 0.85 in Hmean and slower than the baseline experiment.
% --------------------- table begin -----------------------
\begin{table}
  \caption{ Ablation study for the effectiveness of CABs on SCUT-CTW1500.}
  \label{tab:attention}
  \begin{tabular}{ccccc}
    \toprule
      Method &Recall &Precision &Hmean &FPS\\
    \midrule
    baseline  &76.69	&83.86	&80.12 &\textbf{30.21} \\
    with CABs  &\textbf{77.05} 	&\textbf{85.31} 	&\textbf{80.97}  &27.63 \\
  \bottomrule
\end{tabular}
% \vspace{-0.4cm}
\end{table}
% ---------------- table end ---------------------------

\subsection{Evaluation on Curved Text Benchmark}
On SCUT-CTW1500 and Total-Text, we evaluate the performance of SAST for detecting text lines of arbitrary shapes. We fine-tune our model for about 10 epochs on SCUT-CTW1500 and Total-Text training set, respectively. 
In testing phase, the number of vertices of text polygons is adaptively counted and we set the scale of the longer side to 512 for single-scale testing on both datasets.

\begin{figure*}
  \includegraphics[width=\linewidth]{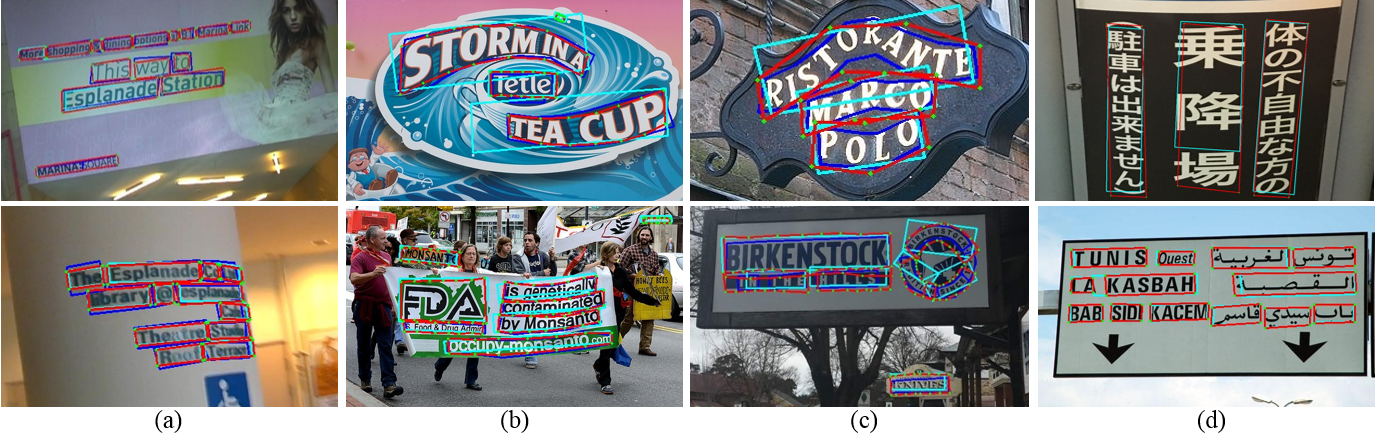}
  \caption{Some qualitative results by the proposed method. From left to right:  ICDAR2015, SCUT-CTW1500, Total-Text, and ICDAR17-MLT. Blue contours: ground truths; Cyan contours: quads from TVO map; Red contours: final detection results. }
  \label{fig:res}
  %\vspace{-0.3cm}
\end{figure*}

%In testing phase, the vertices of text polygons is set to 14 on SCUT-CTW1500, while the number of vertices on Total-Text is adaptive.
The quantitative results are shown in Tab.~\ref{tab:ex_ctw1500} and Tab.~\ref{tab:ex_total_text}. With the help of the efficient post-processing, SAST achieves 80.97\% and 78.08\% in Hmean on SCUT-CTW1500 and Total-Text, respectively, which is comparable to the state-of-the-art methods. In addition, multi-scale testing can further improves Hmean to 81.45\% and 80.21\% on SCUT-CTW1500 and Total-Text. The visualization of curved text detection are shown in Fig.~\ref{fig:res} (a) and (b). As can be seen, the proposed text detector SAST can handle curved text lines well.

%On SCUT-CTW1500 and Total-Text, we evaluate the performance of SAST for detecting text lines of arbitrary shapes. We finetune our model about 10 epochs on SCUT-CTW1500 and Total-Text training set separately. In testing phase, the vertices of text polygons is set to 14 on SCUT-CTW1500, while the number of vertices on Total-Text is adaptive. The quantitative results are shown in Tab. \ref{tab:ex_ctw1500} and Tab. \ref{tab:ex_total_text}. With the help of the accuracy and efficient post-processing, SAST achieves 80.97\% and 78.08\% in Hmean on SCUT-CTW1500 and Total-Text respectively, which is comparable to the state-of-the-art methods. In addition, multi-scale testing can further improves Hmean to 81.45\% and 80.21\% on SCUT-CTW1500 and Total-Text. The visualization of curved text detection are shown in Fig.\ref{fig:res} (a)(b). As can be seen, the proposed text detector SAST can handle well with curved text lines.
% --------------------- table begin -----------------------
\begin{table}
  \caption{Evaluation on SCUT-CTW1500 for detecting text lines of arbitrary shapes.}
  \label{tab:ex_ctw1500}
  \begin{tabular}{ccccc}
    \toprule
      Method & Recall & Precision & Hmean  &FPS \\
    \midrule
    CTPN~\cite{tian2016detecting} &53.80 &60.40 &56.90 &--\\
    EAST~\cite{zhou2017east} &49.10 &78.70 &60.40 &--\\
    DMPNet~\cite{liu2017deep} &56.00 &69.90 &62.20 &--\\
    CTD~\cite{yuliang2017detecting} &65.20 &74.30 &69.50 &15.20\\
    CTD + TLOC~\cite{yuliang2017detecting} &69.80 &74.30 &73.40 &13.30\\
    SLPR~\cite{zhu2018sliding} &70.10 &80.10 &74.80 &--\\
    TextSnake~\cite{long2018textsnake}  &\textbf{85.30} &67.90 &75.60 &12.07\\
    % LOMO &69.62 &89.79 &78.43 \\
    PSENet-4s~\cite{wang2019shape} &78.13 &85.49 &79.29 &8.40 \\
    PSENet-2s~\cite{wang2019shape} &79.30 &81.95 &80.60 &-- \\
    % LOMO MS &76.53 &85.66 &80.84 \\
    PSENet-1s~\cite{wang2019shape} &79.89 &82.50 &81.17 &3.90\\
    TextField~\cite{xu2019textfield} &79.80 &83.00 &81.40 &--\\
%    TextMountain &83.40 &82.90 &83.20 \\
    \midrule
    SAST  &77.05 	&\textbf{85.31} 	&80.97  &27.63\\
    SAST MS &81.71 	&81.19 	&\textbf{81.45}  &--\\

  \bottomrule
\end{tabular}
% \vspace{-0.4cm}
\end{table}
% ---------------- table end ---------------------------

% --------------------- table begin -----------------------
\begin{table}
  \caption{Evaluation on Total-Text for detecting text lines of arbitrary shapes.}
  \label{tab:ex_total_text}
  \begin{tabular}{cccc}
    \toprule
      Method & Recall & Precision & Hmean\\
    \midrule
    SegLink~\cite{shi2017detecting} &23.80 &30.30 &26.70 \\
    DeconvNet~\cite{ch2017total} &33.00 &40.00 &36.00 \\
    EAST~\cite{zhou2017east} &36.20 &50.00 &42.00 \\
    TextSnake~\cite{long2018textsnake} &74.50 &82.70 &78.40 \\
    TextField~\cite{xu2019textfield} &\textbf{79.90} &81.20 &\textbf{80.60} \\
    % LOMO &75.70 &88.60 &81.60 \\
    % SPCNet &82.80 &83.00 &82.90 \\
    % LOMO MS &79.30 &87.60 &83.30 \\

    \midrule
    SAST     &76.86 	&83.77 	&80.17  \\
    SAST MS  &75.49 	&\textbf{85.57} 	&80.21  \\

  \bottomrule
\end{tabular}
%\vspace{-0.4cm}
\end{table}
% ---------------- table end ---------------------------

% ------------------- figure begin ----------------
% ------------------- figure end ------------------

\subsection{Evaluation on ICDAR 2015}
In order to verify the validity for detecting oriented text, we compare SAST with the state-of-the-art methods on ICDAR 2015 dataset, a standard oriented text dataset. 
%We set the scale of the longer side to 1536 and keep the aspect ratio unchanged for single-scale testing. Longer sides in multi-scale testing are set to 1024, 1536, and 2048.
Compared with previous arbitrarily-shaped text detectors~\cite{xu2019textfield, yang2018inceptext,wang2019shape}, which detect text on the same size as input image, SAST achieves a better performance in a much faster speed. All the results are listed in Tab.~\ref{tab:ex_icdar15}. Specifically, for single-scale testing, SAST achieves 86.91\% in Hmean, surpassing most competitors (these pure detection methods without the assistance of recognition task). Moreover, multi-scale testing increases about 0.53\% in Hmean. Some detection results are shown in Fig.~\ref{fig:res} (c), and indicate that SAST is also capable to detect multi-oriented text accurately.
% --------------------- table begin -----------------------
\begin{table}
  \caption{ Evaluation on ICDAR 2015 for detecting oriented text.}
  \label{tab:ex_icdar15}
  \begin{tabular}{cccc}
    \toprule
      Method & Recall & Precision & Hmean\\
    \midrule
    DMPNet~\cite{liu2017deep} &68.22 &73.23 &70.64 \\
    SegLink~\cite{shi2017detecting} &76.50 &74.74 &75.61 \\
    %MCN &80.00 &72.00 &76.00 \\
    SSTD~\cite{he2017single} &73.86 &80.23 &76.91 \\
    WordSup~\cite{hu2017wordsup} &77.03 &79.33 &78.16 \\
    %ITN &74.10 &85.70 &79.50 \\
    RRPN~\cite{ma2018arbitrary} &77.13 &83.52 &80.20 \\
    EAST~\cite{zhou2017east} &78.33 &83.27 &80.72 \\
    He et al.~\cite{he2017deep} &80.00 &82.00 &81.00 \\
    TextField~\cite{xu2019textfield}&80.50 &84.30 &82.40 \\
    %R2CNN &79.68 &85.62 &82.54 \\
    TextSnake~\cite{long2018textsnake} &80.40 &84.90 &82.60 \\
    PixelLink~\cite{deng2018pixellink} &82.00 &85.50 &83.70 \\
    RRD~\cite{liao2018rotation}&80.00 &88.00 &83.80 \\
    %CCFLAB\_FTSN &80.07 &88.65 &84.14 \\
    Lyu et al.~\cite{lyu2018multi} &79.70 &\textbf{89.50} &84.30 \\
    %SLPR &83.60 &85.50 &84.50 \\
    PSENet-4s~\cite{wang2019shape} &83.87 &87.98 &85.88 \\
    %Mask TextSpotter\cite{yao2018mask} &81.00 &91.60 &86.00 \\
%    TextMountain &84.16 &88.51 &86.28 \\
    IncepText~\cite{yang2018inceptext} &84.30 &89.40 &86.80 \\
%    End-to-End TextSpotter\cite{buvsta2017deep} &86.00 &87.00 &87.00 \\
    PSENet-1s~\cite{wang2019shape} &85.51 &88.71 &87.08 \\
%    SPCNet &85.80 &88.70 &87.20 \\
%    LOMO &83.49 &91.26 &87.20 \\
    PSENet-2s~\cite{wang2019shape} &85.22 &89.30 &87.21 \\
%    TextNet &85.41 &89.42 &87.37 \\
%    Pixel-Anchor &87.05 &88.32 &87.68 \\
%    LOMO MS &87.63 &87.84 &87.73 \\
    \midrule
    %SAST     &85.98 &88.37 	& 87.16  \\
    %SAST MS  &\textbf{88.25} &86.26 	& \textbf{87.24} \\
    SAST     &87.09 &86.72 	& 86.91  \\
    SAST MS  &\textbf{87.34} &87.55 	& \textbf{87.44} \\
  \bottomrule
\end{tabular}
%\vspace{-0.4cm}
\end{table}

\subsection{Evaluation on ICDAR2017-MLT}
To demonstrate the generalization ability of SAST on multilingual scene text detection, we evaluate SAST on ICDAR2017-MLT. Similar to the above training methods, the detector is fine-tuned for about 10 epochs on the SynthText pre-trained model. At the single scale testing, our proposed method achieves a Hmean of 68.76\%, and it increases to 72.37\% for multi-scale testing. The quantitative results are shown in Tab.~\ref{tab:ex_icdar17mlt}. The visualizatio n of multilingual text detection is as illustrated in the Fig.~\ref{fig:res} (d), which shows the robustness of the proposed method in detecting multilingual scene text.

% , which obtains state-of-the-art performance for in this dataset.

% --------------------- table begin -----------------------
\begin{table}
  \caption{Evaluation on ICDAR2017-MLT for the generalization ability of SAST on multilingual scene text detection.}
  \label{tab:ex_icdar17mlt}
  \begin{tabular}{cccc}
    \toprule
      Method & Recall & Precision & Hmean\\
    \midrule
    %SCUT DLVClab &54.50 &80.30 &65.00 \\
    %NLPR PAL &57.90 &76.70 &66.00 \\
    Lyu et al.~\cite{lyu2018multi} &56.60 &\textbf{83.80} &66.80 \\
    % Pixel-Anchor &59.54 &79.54 &68.10 \\
    % LOMO &60.56 &78.77 &68.47 \\
    AF-RPN~\cite{zhong2018anchor} &66.00 &75.00 &70.00 \\
    % SPCNet &66.90 &73.40 &70.00 \\
    PSENet-4s~\cite{wang2019shape} &67.56 &75.98 &71.52 \\
    PSENet-2s~\cite{wang2019shape} &68.35 &76.97 &72.40 \\
    Lyu et al. MS~\cite{lyu2018multi} &70.60 &74.30 &72.40 \\
    PSENet-1s~\cite{wang2019shape} &\textbf{68.40} &77.01 &\textbf{72.45} \\
    % LOMO MS &67.62 &79.59 &73.12 \\
    % SPCNet MS &68.60 &80.60 &74.10 \\
    \midrule
    SAST     &67.56 	&70.00 	&68.76 \\
    SAST MS &66.53 	&79.35 	&72.37 \\
  \bottomrule
\end{tabular}
%\vspace{-0.4cm}
\end{table}
% ---------------- table end ---------------------------

\subsection{Runtime}
In this paper, we make a trade-off between speed and accuracy. The TCL, TVO, TCO, and TBO maps are predicted in the 1/4 size of input images. With the proposed post-processing step, SAST is supposed to detect text of arbitrary shapes in real-time speed with a commonly used GPU. To demonstrate the runtime of the proposed method, we run testing on SCUT-CTW1500 with a workstation equipped with NVIDIA TITAN Xp. The test image is resized to $512 \times 512$ and the batch size is set to 1 on a single GPU. It takes 29.58 ms and 6.61 ms in the process of network inference and post-processing respectively, which is written in Python code \footnote{The NMS part is written in C++.} and can be further optimized. It runs at 27.63 FPS with a Hmean of 80.97\%, surpassing most of the existing arbitrarily-shaped text detectors in both accuracy and efficiency\footnote{The speed of different methods is depicted for reference only, which might be evaluated with different hardware environments.}, as depicted in Tab.~\ref{tab:ex_ctw1500}. 

% More details are shown in the Tab. \ref{tab:ex_runtime}. 
% Meanwhile, we also test our runtime on ICDAR15 dataset to compare with the most previous works more fairly, as list in Tab. \ref{tab:ex_icdar15}. 

% % --------------------- table begin -----------------------
% \begin{table}
%   \caption{Evaluation on SCUT-CTW1500 for runtime.}
%   \label{tab:ex_runtime}
%   \begin{tabular}{ccccc}
%     \toprule
%       Device & Max size  & $T_{infer}$ & $T_{post}$ & FPS \\
%     \midrule
%     % Nvidia Tesla K40&    512     &82.43   & 6.72 &11.22\\
%     Nvidia TITAN Xp	&   512   	&29.58	&6.61  &27.63 \\ 
%   \bottomrule
% \end{tabular}
% \end{table}
% % ---------------- table end ---------------------------

% % --------------------- table begin -----------------------
% \begin{table}
%   \caption{Evaluation on on SCUT-CTW1500 for runtime}
%   \label{tab:ex_runtime}
%   \begin{tabular}{ccccc}
%     \toprule
%      NO. &  Max size  & $T_{infer}(ms)$ & $T_{post}(ms)$ & FPS \\
%     \midrule
%     1&  512 & \\
%     2&  512 & \\
%     3&  \\
%     \midrule
%     avg. & \\
%     ratio. & \\
%   \bottomrule
% \end{tabular}
% \end{table}
% % ---------------- table end ---------------------------

\section{Conclusion and Future Work}
In this paper, we propose an efficient single-shot arbitrarily-shaped text detector together with Context Attention Blocks and a mechanism of point-to-quad assignment, which integrates both high-level object knowledge and low-level pixel information to obtain text instances from a context-enhanced segmentation. Several experiments demonstrate that the proposed SAST is effective in detecting arbitrarily-shaped text, and is also robust in generalizing to multilingual scene text datasets. Qualitative results show that SAST helps to alleviate some common challenges in segmentation-based text detector, such as the problem of fragments and the separation of adjacent text instances. Moreover, with a commonly used GPU, SAST runs fast and may be sufficient for some real-time applications, e.g., augmented reality translation. However, it is difficult for SAST to detect some extreme cases, which mainly are very small text regions. In the future, we are interested in improving the ability of small text detection and developing an end-to-end text reading system for text of arbitrary shapes.

% \begin{table}
%   \caption{}
%   \label{tab:freq}
%   \begin{tabular}{ccl}
%     \toprule
%     Non-English or Math&Frequency&Comments\\
%     \midrule
%     \O & 1 in 1,000& For Swedish names\\
%     $\pi$ & 1 in 5& Common in math\\
%     \$ & 4 in 5 & Used in business\\
%     $\Psi^2_1$ & 1 in 40,000& Unexplained usage\\
%   \bottomrule
% \end{tabular}
% \end{table}

% \section*{Acknowledgements}
\begin{acks}
  This work is supported in part by the National Natural Science Foundation of
  China (NSFC) under Grant 61572387, Grant 61632019, Grant 61836008,
  and Grant 61672404, and the Foundation for Innovative Research
  Groups of the NSFC under Grant 61621005.
\end{acks}

%
% The next two lines define the bibliography style to be used, and the bibliography file.
\bibliographystyle{ACM-Reference-Format}
\bibliography{sast_updated}

% 
% If your work has an appendix, this is the place to put it.
%\appendix

\end{document}